\documentclass[letterpaper]{article} 
\usepackage{aaai23}  
\usepackage{times}  
\usepackage{helvet}  
\usepackage{courier}  
\usepackage[hyphens]{url}  
\usepackage{graphicx} 
\usepackage{natbib}  
\usepackage{caption} 
\usepackage{algorithm}
\usepackage{algorithmic}
\usepackage{overpic}
\usepackage{rotating}
\usepackage{multirow}
\usepackage{makecell}
\usepackage{todonotes}
\usepackage{amsmath}
\usepackage[capitalize]{cleveref}
\usepackage{amssymb}
\usepackage{booktabs}
\usepackage{newfloat}
\usepackage{listings}
\DeclareCaptionStyle{ruled}{labelfont=normalfont,labelsep=colon,strut=off} 
\floatstyle{ruled}
\newfloat{listing}{tb}{lst}{}
\floatname{listing}{Listing}
\newcommand{\blue}[1]{\textcolor{blue}{#1}}

\def\eg{\emph{e.g.}}

\def\ourmodel{AFFormer}

\title{Head-Free Lightweight Semantic Segmentation with Linear Transformer}

\author {
    Bo Dong \textsuperscript{\rm 1}\thanks{Work done during an internship at Alibaba Group.},
    Pichao Wang \textsuperscript{\rm 1 }\thanks{Corresponding author; work done at Alibaba Group, and now affiliated with Amazon Prime Video.},
    Fan Wang \textsuperscript{\rm 2}
}
\affiliations {
    \textsuperscript{\rm 1} Alibaba Group\\
    \{bo.dong.cst, pichaowang\}@gmail.com; fan.w@alibaba-inc.com
}

\usepackage{bibentry}

\begin{document}

\maketitle

\begin{abstract}
Existing semantic segmentation works have been mainly focused on designing effective decoders; however, the computational load introduced by the overall structure has long been ignored, which hinders their applications on resource-constrained hardwares. 
In this paper, we propose a head-free lightweight architecture specifically for semantic segmentation, named Adaptive Frequency Transformer~(\ourmodel).
\ourmodel~adopts a parallel architecture to leverage prototype representations as specific learnable local descriptions which replaces the decoder and preserves the rich image semantics on high-resolution features.
Although removing the decoder compresses most of the computation, the accuracy of the parallel structure is still limited by low computational resources.
Therefore, we employ heterogeneous operators (CNN and Vision Transformer) for pixel embedding and prototype representations to further save computational costs. 
Moreover, it is very difficult to linearize the complexity of the vision Transformer from the perspective of spatial domain. 
Due to the fact that semantic segmentation is very sensitive to frequency information, we construct a lightweight prototype learning block with adaptive frequency filter of complexity $O(n)$ to replace standard self attention with $O(n^{2})$.
Extensive experiments on widely adopted datasets demonstrate that \ourmodel~achieves superior accuracy while retaining only 3M parameters.
On the ADE20K dataset, \ourmodel~achieves 41.8 mIoU and 4.6 GFLOPs, which is 4.4 mIoU higher than Segformer, with  45\% less GFLOPs. On the Cityscapes dataset, \ourmodel~achieves 78.7 mIoU and 34.4 GFLOPs, which is 2.5 mIoU higher than Segformer with 72.5\% less GFLOPs. Code is available at https://github.com/dongbo811/AFFormer.
\end{abstract}

\section{Introduction}
\label{sec:intro}
\setlength{\abovecaptionskip}{1.5pt}%
\setlength{\belowcaptionskip}{1.5pt}%
\begin{figure}[t!]
	\centering
    \small
	\begin{overpic}[width=1.\linewidth]{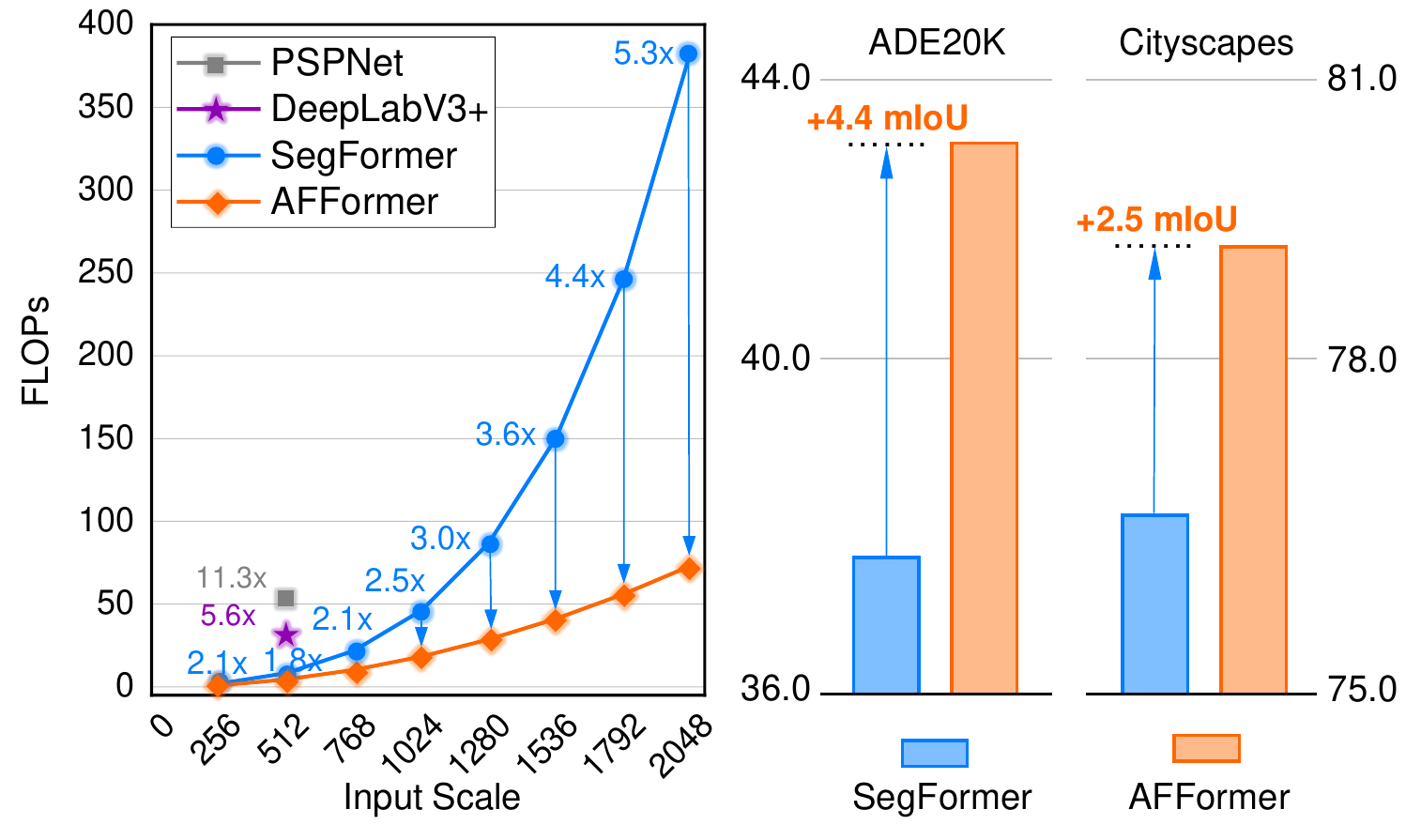}
    \end{overpic}
	\caption{\textit{Left:} Computational complexity under different input scales. Segformer~\cite{xie2021segformer} significantly reduces the computational complexity compared to traditional methods, such as PSPNet~\cite{zhao2017pyramid} and DeepLabV3+~\cite{chen2018encoder} which have mobilenetV2~\cite{sandler2018mobilenetv2} as backbone. However, Segformer still has a huge computational burden for higher resolutions. \textit{Right:} \ourmodel~achieves better accuracy on ADE20K and Cityscapes datasets with significantly lower FLOPs.}
    \label{figure: flops}
\end{figure}

Semantic segmentation aims to partition an image into sub-regions (collections of pixels) and is defined as a pixel-level classification task~\cite{long2015fully,xie2021segformer,zhao2017pyramid,chen2018encoder,strudel2021segmenter,cheng2021maskformer} since Fully Convolutional Networks (FCN)~\cite{long2015fully}.
It has two unique characteristics compared to image classification: pixel-wise dense prediction and multi-class representation, which is usually built upon high-resolution features and requires a global inductive capability of image semantics, respectively.
Previous semantic segmentation methods~\cite{zhao2017pyramid, chen2018encoder,strudel2021segmenter,xie2021segformer,cheng2021maskformer,yuan2021ocnet} focus on using the classification network as backbone to extract multi-scale features, and designing a complicated decoder head to establish the relationship between multi-scale features. 
However, these improvements come at the expense of large model size and high computational cost.
For instance, the well-known PSPNet~\cite{zhao2017pyramid} using light-weight MobilenetV2~\cite{sandler2018mobilenetv2} as backbone contains 13.7M parameters and 52.2 GFLOPs with the input scale of $512 \times 512$.
The widely-used DeepLabV3+~\cite{chen2018encoder} with the same backbone requires 15.4M parameters and 25.8 GFLOPs. 
The inherent design manner limits the development of this field and hinders many real-world applications. Thus, we raise the following question: \textit{can semantic segmentation be as simple as image classification?}

Recently vision Transformers (ViTs)~\cite{liu2021swin,lee2022mpvit, xie2021segformer,strudel2021segmenter,cheng2021maskformer,Xu_2021_ICCV,lee2022mpvit} have shown great potential in semantic segmentation, however, they face the challenges of balancing performance and memory usage when deployed on ultra-low computing power devices.
Standard Transformers has computational complexity of $O(n^{2})$ in the spatial domain, where $n$ is the input resolution. 
Existing methods alleviate this situation by reducing the number of tokens~\cite{xie2021segformer,wang2021pvt,liang2022evit,ren2022shunted} or sliding windows~\cite{liu2021swin,yuan2021hrformer}, but they introduce limited  reduction on computational complexity and even compromise global or local semantics for the segmentation task. 
Meanwhile, semantic segmentation as a fundamental research field, has extensive application scenarios and needs to process images with various resolutions. 
As shown in Figure~\ref{figure: flops}, although the well-known efficient Segformer~\cite{xie2021segformer} achieves a great breakthrough compared to PSPNet and DeepLabV3+, it still faces a huge computational burden for higher resolutions. 
At the scale of $512 \times 512$, although Segformer is very light compared to PSPNet and DeepLabV3+, it is almost twice as expensive as ours (8.4 GFLOPs \textit{vs} 4.6 GFLOPs); at the scale of $2048 \times 2048$, even 5x GFLOPs is required (384.3 GFLOPs \textit{vs} 73.2 GFLOPs). 
Thus, we raise another question: \textit{can we design an efficient and lightweight Transformer network for semantic segmentation in ultra-low computational scenarios?}

The answers to above two questions are affirmative. To this end, we propose a head-free lightweight semantic segmentation specific architecture, named Adaptive Frequency Transformer~(\ourmodel). 
Inspired by the properties that ViT maintains a single high-resolution feature map to keep details~\cite{dosovitskiy2020vit} and the pyramid structure reduces the resolution to explore semantics and reduce computational cost~\cite{he2016resnet,wang2021pvt,liu2021swin}, \ourmodel~adopts a parallel architecture to leverage the prototype representations as specific learnable local descriptions which replace the decoder and preserves the rich image semantics on high-resolution features.
The parallel structure compresses the majority of the computation by removing the decoder, but it is still not enough for ultra-low computational resources. Moreover, we employ heterogeneous operators for pixel embedding features and local description features to save more computational costs. 
A Transformer-based module named \textit{prototype learning} (PL) is used to learn the prototype representations, while a convolution-based module called \textit{pixel descriptor} (PD) takes pixel embedding features and the learned prototype representations as inputs, transforming them back into the full pixel embedding space to preserve high-resolution semantics.

However, it is still very difficult to linearize the complexity of the vision Transformer from the perspective of spatial domain. 
Inspired by the effects of frequency on classification tasks~\cite{rao2021global,wang2020high}, we find that semantic segmentation is also very sensitive to frequency information. 
Thus, we construct a lightweight adaptive frequency filter of complexity $O(n)$ as prototype learning to replace the standard self attention with $O(n^{2})$.
The core of this module is composed of frequency similarity kernel, dynamic low-pass and high-pass filters, which capture frequency information that is beneficial to semantic segmentation from the perspectives of emphasizing important frequency components and dynamically filtering frequency, respectively.
Finally, the computational cost is further reduced by sharing weights in high and low frequency extraction and enhancement modules.
We also embed a simplified depthwise convolutional layer in the feed-forward network (FFN) layer to enhance the fusion effect, reducing the size of the two matrix transformations.

With the help of parallel heterogeneous architecture and adaptive frequency filter, we use only one convolutional layer as classification layer (CLS) for single-scale feature, achieving the best performance and making semantic segmentation as simple as image classification. 
We demonstrate the advantages of the proposed \ourmodel~on three widely-used datasets: ADE20K, Cityscapes and COCO-stuff.
With only 3M parameters, \ourmodel~significantly outperforms the state-of-the-art lightweight methods. 
On ADE20K, \ourmodel~achieves 41.8 mIoU with 4.6 GFLOPs, outperforming Segformer by 4.4 mIoU, while reducing GFLOPs by 45\%. 
On Cityscapes, \ourmodel~achieves 78.7 mIoU and 34.4 GFLOPs, which is 2.5 mIoU higher than Segformer, with 72.5\% less GFLOPs.
Extensive experimental results demonstrate that it is possible to apply our model in computationally constrained scenarios, which still maintaining the high performance and robustness across different datasets.

\begin{figure*}[t!]
	\centering
    \small
	\begin{overpic}[width=1.\linewidth]{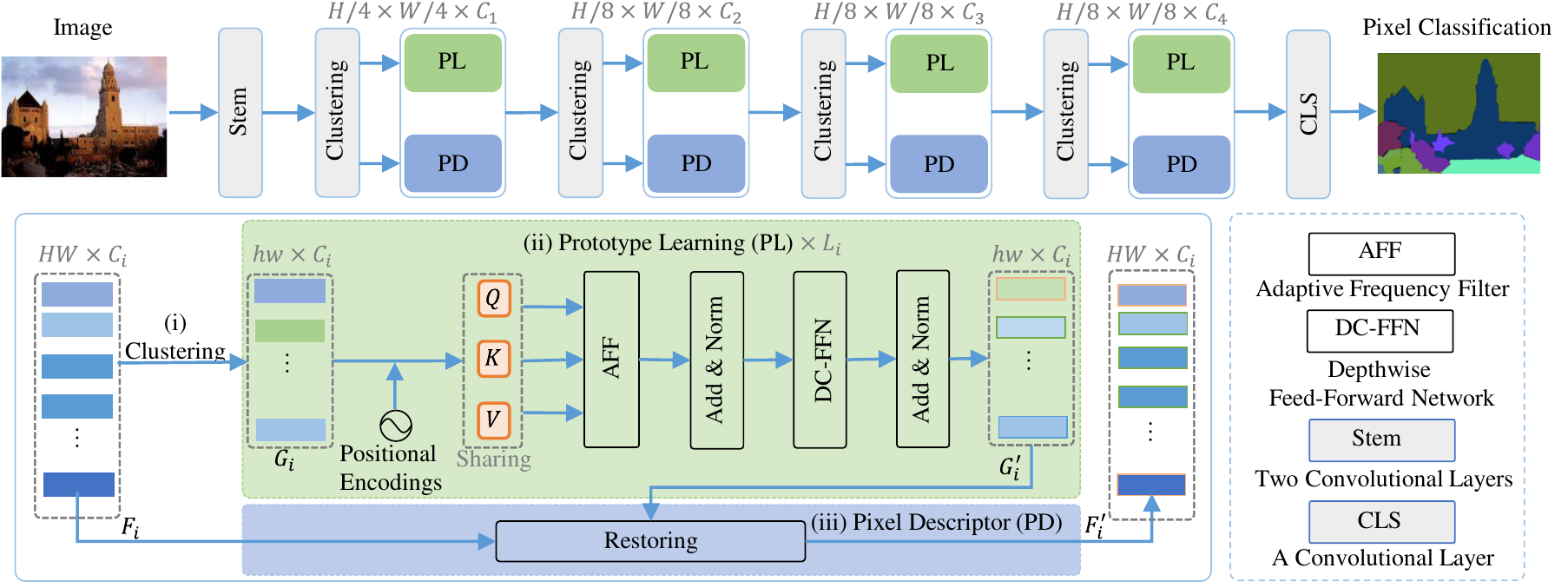}
    \end{overpic}
\caption{An Overview of Adaptive Frequency Transformer~(\ourmodel). We first displays the overall structure of parallel heterogeneous network.
Specifically, the feature $\boldsymbol{F}$ after patch embedding is first clustered to obtain the prototype feature $\boldsymbol{G}$, so as to construct a parallel network structure, which includes two heterogeneous operators. 
A Transformer-based module as prototype learning to capture favorable frequency components in $\boldsymbol{G}$, resulting prototype representation $\boldsymbol{G}^{\prime}$.
Finally $\boldsymbol{G}^{\prime}$ is restored by a CNN-based pixel descriptor, resulting $\boldsymbol{F}^{\prime}$ for the next stage.}
    \label{figure: parallel}
\end{figure*}

\section{Related Work}
\label{sec:Related Work}

\subsection{Semantic Segmentation}
Semantic segmentation is regarded as a pixel classification task~\cite{strudel2021segmenter,xu2017deep,xie2021segformer}. 
In the last two years, new paradigms based on visual Transformers have emerged, which enable mask classification via queries or dynamic kernels~\cite{zhang2021knet,li2022maskdino,cheng2021maskformer,cheng2022mask2former}. 
For instance, Maskformer~\cite{cheng2021maskformer} learns an object query and converts it into an embedding of masks. 
Mask2former~\cite{cheng2022mask2former} enhances the query learning with a powerful multi-scale masked Transformer~\cite{zhu2020deformable}. 
K-Net~\cite{zhang2021knet} adopts dynamic kernels for masks generation. 
MaskDINO~\cite{li2022maskdino} brings object detection to semantic segmentation, further improving query capabilities.
However, all above methods are not suitable for low computing power scene due to the high computational cost of learning efficient queries and dynamic kernels.
We argue that the essence of these paradigms is to update pixel semantics by replacing the whole with individual representations.
Therefore, we leverage pixel embeddings as a specific learnable local description that extracts image and pixel semantics and allows semantic interaction.

\subsection{Efficient Vision Transformers}
The lightweight solution of vision Transformer mainly focuses on the optimization of self attention, including following ways: reducing the token length~\cite{wang2021pvt, xie2021segformer,wang2022pvtv2} and using local windows~\cite{liu2021swin,yuan2021hrformer}.
PVT~\cite{wang2021pvt} performs spatial compression on keys and values through spatial reduction, and PVTv2~\cite{wang2022pvtv2} further replaces the spatial reduction by pooling operation, but many details are lost in this way.
Swin~\cite{liu2021swin,yuan2021hrformer} significantly reduce the length of the token by restricting self attention to local windows, while these against the global nature of Transformer and restrict the global receptive field. 
At the same time, many lightweight designs~\cite{chen2022mobileformer,mehta2022mobilevit} introduce Transformers in MobileNet to obtain more global semantics, but these methods still suffer from the square-level computational complexity of conventional Transformers.
Mobile-Former~\cite{chen2022mobileformer} combines the parallel design of MobileNet~\cite{sandler2018mobilenetv2} and Transformer~\cite{dosovitskiy2020vit}, which can achieve bidirectional fusion performance of local and global features far beyond lightweight networks such as MobileNetV3. However, it only uses a very small number of tokens, which is not conducive to semantic segmentation tasks.

\section{Method}\label{sec:Method_Analysis}

In this section, we introduce the lightweight parallel heterogeneous network for semantic segmentation. 
The basic information is first provivided on the replacement of semantic decoder by parallel heterogeneous network. 
Then, we introduce the modeling of pixel descriptions and semantic frequencies. Finally, the specific details and the computational overhead of parallel architectures are discussed.

\subsection{Parallel Heterogeneous Architecture}
\label{subsec:HPA}

The semantic decoder propagates the image semantics obtained by the encoder to each pixel and restores the lost details in downsampling. 
A straightforward alternative is to extract image semantics in high resolution features, but it introduces a huge amount of computation, especially for vision Transformers. In contrast, we propose a novel strategy to describe pixel semantic information with prototype semantics. For each stage, given a feature $\boldsymbol{F} \in \mathbb{R}^{H \times W \times C} $, we first initial a grid $\boldsymbol{G} \in \mathbb{R}^{h \times w \times C} $ as a prototype of the image, where each point in $\boldsymbol{G}$ acts as a local cluster center, and the initial state simply contains information about the surrounding area. Here we use a $1 \times C$ vector to represent the local semantic information of each point. For each specific pixel, because the semantics of the surrounding pixels are not consistent, there are overlap semantics between each cluster centers. The cluster centers are weighted initialized in its corresponding area $\alpha^{2}$, and the initialization of each cluster center is expressed as:
\begin{equation}
\boldsymbol{G}(s) = \sum_{i=0}^{n}w_{i}x_{i} 
\end{equation} 
where $n=\alpha \times \alpha$, $w_{i}$ denotes the weight of $x_{i}$, and $\alpha$ is set to 3. Our purpose is to update each cluster center $s$ in the grid $G$ instead of updating the feature $\boldsymbol{F}$ directly. As $h \times w \ll H\times W$, it greatly simplifies the computation.

Here, we use a Transformer-based module as prototype learning to update each cluster center, which contains $L$ layers in total, and the updated center is denoted as $\boldsymbol{G}^{\prime}(s)$. For each updated cluster center, we recover it by a pixel descriptor. Let $F_{i}^{\prime}$ denote the recovered feature, which contains not only the rich pixel semantics from $F$, but also the prototype semantics collected by the cluster centers $\boldsymbol{G}^{\prime}(s)$. 
Since the cluster centers aggregate the semantics of surrounding pixels, resulting in the loss of local details, PD first models local details in $F$ with pixel semantics. 
Specifically, $F$ is projected to a low-dimensional space, establishing local relationships between pixels such that each local patch keeps a distinct boundary. 
Then $\boldsymbol{G}^{\prime}(s)$ is embedded into $F$ to restore to the original space feature $F^{\prime}$ through bilinear interpolation. 
Finally, they are integrated through a linear projection layer. 

\begin{figure}[t!]
	\centering
    \small
	\begin{overpic}[width=.9\linewidth]{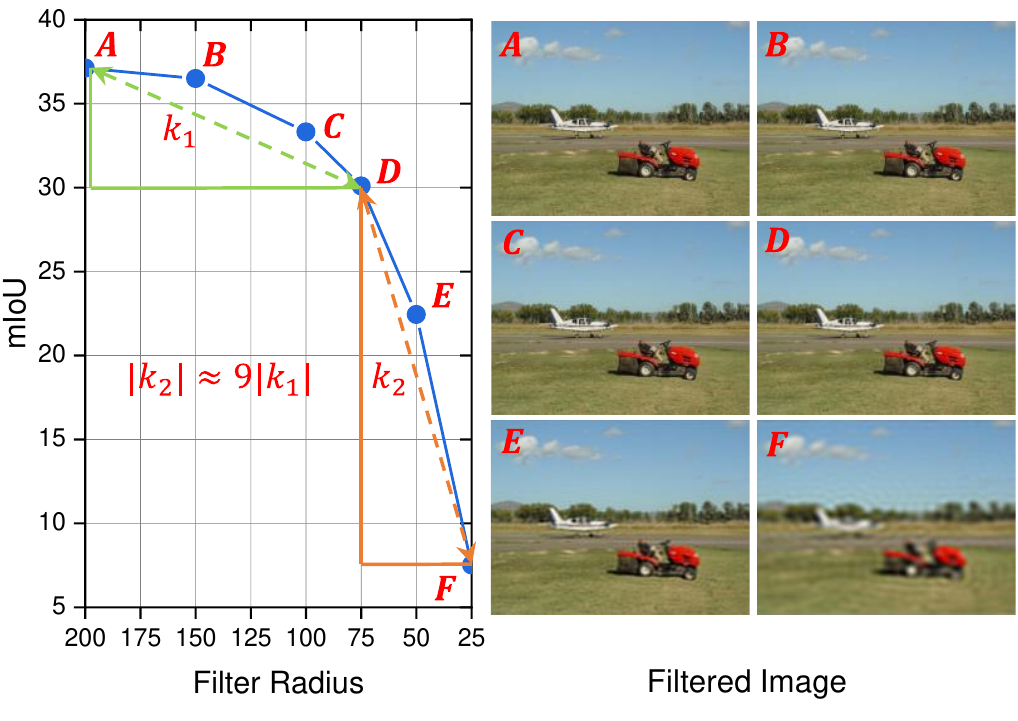}
    \end{overpic}
	\caption{The effect of different frequency components on semantic segmentation. We use the cut-edge method Segformer~\cite{xie2021segformer} to evaluate the impact of frequency components on semantic segmentation on the widely used ADE20K dataset~\cite{zhou2017scene}.
The image is transformed into the frequency domain by a fast Fourier transform ~\cite{heideman1984gauss}, and high-frequency information is filtered out using a low-pass operator with a radius.
Removing high-frequency components at different levels results the prediction performance drops significantly.}
    \label{figure: filter}
\end{figure}

\subsection{Prototype Learning by Adaptive Frequency Filter}\label{subsec:Frequency Aware Statistics}
\subsubsection{Motivation}
Semantic segmentation is an extremely complex pixel-level classification task that is prone to category confusion.
The frequency representation can be used as a new paradigm of learning difference between categories, which can excavate the information ignored by human vision~\cite{zhong2022detecting,qian2020thinking}. 
As shown in Figure~\ref{figure: filter}, humans are robust to frequency information removal unless the vast majority of frequency components are filtered out. 
However, the model is extremely sensitive to frequency information removal, and even removing a small amount would result in significant performance degradation.
It shows that for the model, mining more frequency information can enhance the difference between categories and make the boundary between each category more clear, thereby improving the effect of semantic segmentation.

\begin{figure}[t!]
	\centering
    \small
	\begin{overpic}[width=.9\linewidth]{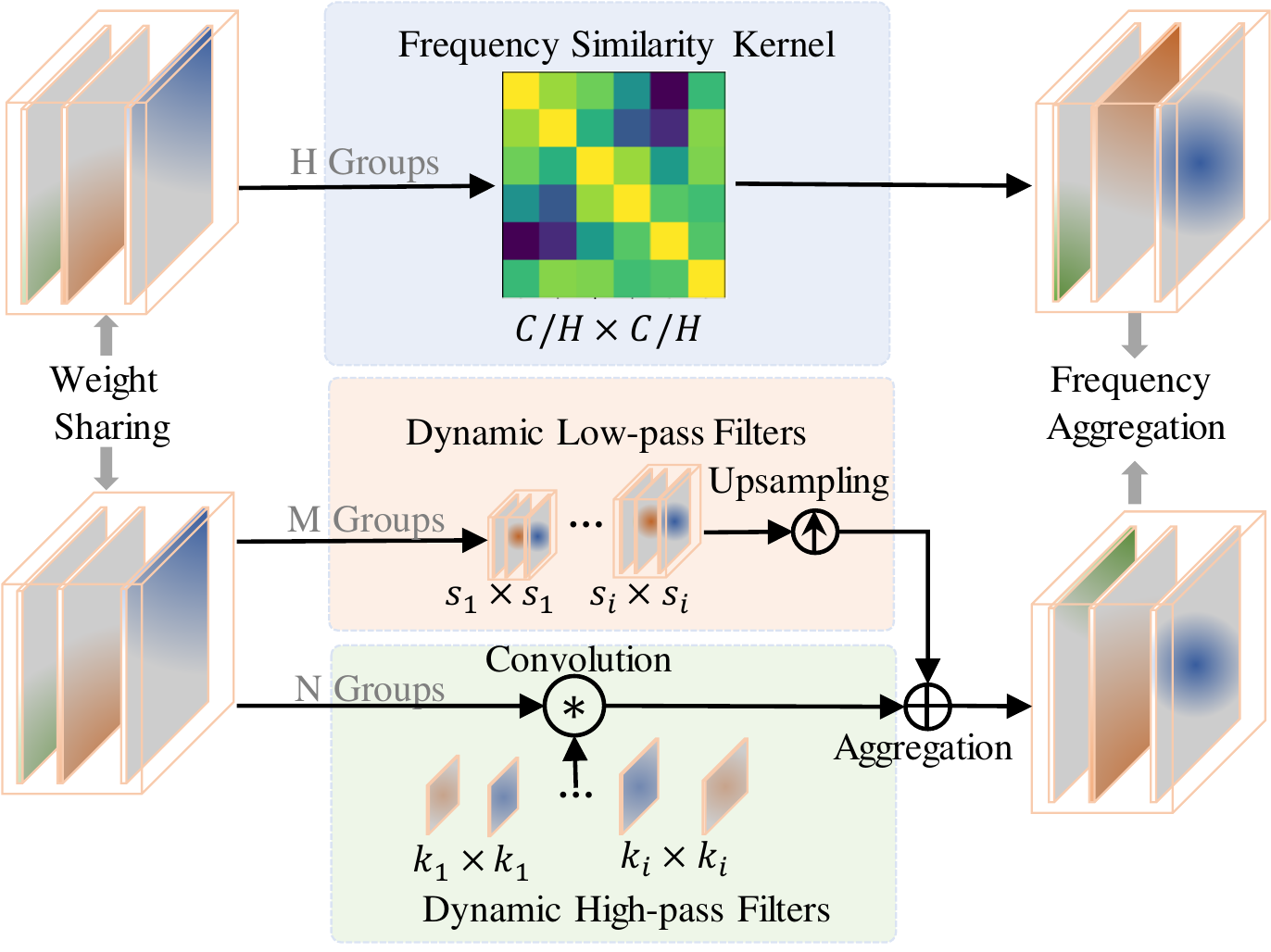}
    \end{overpic}
	\caption{Structure of the adaptive frequency filter in prototype learning. The prototype as learnable local description utilizes frequency component similarity kernel to enhance different components while combining efficient and dynamic low-pass and high-pass filters to capture more frequency information.}
    \label{figure: AFF}
\end{figure}

Since feature $\boldsymbol{F}$ contains rich frequency features, each cluster center in the grid $\boldsymbol{G}$ also collects these frequency information.
Motivated by the above analysis, extracting more beneficial frequencies in grid $\boldsymbol{G}$ helps to discriminate the attributes of each cluster.
To extract different frequency features, the straightforward way is to transform the spatial domain features into spectral features through Fourier transform, and use a simple mask filter in the frequency domain to enhance or attenuate the intensity of each frequency component of the spectrum. 
Then the extracted frequency features are converted to the spatial domain by inverse Fourier transform. 
However, Fourier transform and inverse transform bring in additional computational expenses, and such operators are not supported on many hardwares. 
Thus, we design an adaptive frequency filter block based on the vanilla vision Transformer from the perspective of spectral correlation to capture important high frequency and low frequency features directly in the spatial domain. 
The core components are shown in Figure~\ref{figure: AFF} and the formula is defined as:
\setlength{\abovedisplayskip}{10pt}
\begin{equation}
  \boldsymbol{AFF}(X) = \underbrace{||\boldsymbol{D}_{h}^{fc}(X)||_{H}}%
    _{corr.} + \underbrace{||\boldsymbol{D}_{m}^{lf}(X)||_{M} + ||\boldsymbol{D}_{n}^{hf}(X)||_{N}}
    _{dynamic~filters},
\end{equation}
\setlength{\belowdisplayskip}{10pt}
where $\boldsymbol{D}_{h}^{fc}$, $\boldsymbol{D}_{m}^{lf}(X)$ and $\boldsymbol{D}_{n}^{hf}(X)$ denote the frequency similarity kernel with ${H}$ groups to achieve frequency component correlation enhancement, dynamical low-pass filters with ${M}$ groups and dynamical high-pass filters with ${N}$ groups, respectively. $||\cdot||$ denotes concatenation.
It is worth noting that these operators adopt a parallel structure to further reduce the computational cost by sharing weights.

\subsubsection{Frequency Similarity Kernel (FSK)}
Different frequency components distribute over in $\boldsymbol{G}$, and our purpose is to select and enhance the important components that helps semantic parsing. 
To this end, we design a frequency similarity kernel module. Generally, this module is implemented by the vision Transformer. 
Given a feature $\boldsymbol{X}\in \mathbb{R}^{ (hw)\times C }$, with relative position encoding on $\boldsymbol{G}$ through a convolution layer~\cite{Wu2021RPE}. 
We first use a fixed-size similarity kernel $\boldsymbol{A} \in \mathbb{R}^{C/H \times C/H }$ to represent the correspondence between different frequency components, and select the important frequency components by querying the similarity kernel.
We treat it as a function transfer that computes the keys $\boldsymbol{K}$ and values $\boldsymbol{V}$ of frequency components through a linear layer, and normalizes the keys across frequency components by a Softmax operation. Each component integrates a similarity kernel $\boldsymbol{A}_{i,j}$, which is computed as:

\begin{equation}
\boldsymbol{A}_{i,j} = { e^{\boldsymbol{k}_i\boldsymbol{v}_j^{\top}}}/{\sum\limits_{j=1}^n e^{\boldsymbol{k}_i}},
\label{eq:std-att-2}
\end{equation}
where $\boldsymbol{k}_i$ represents the $i$-th frequency component in $\boldsymbol{{K}}$, $\boldsymbol{v}_j$ represents the $j$-th frequency component in $\boldsymbol{V}$. We also transform the input $\boldsymbol{X}$ into the query $\boldsymbol{Q}$ through a linear layer, and obtain the component-enhanced output through interactions on the fixed-size similarity kernel.

\subsubsection{Dynamic Low-Pass Filters (DLF)}
Low-frequency components occupy most of the energy in the absolute image and represent most of the semantic information. 
A low-pass filter allows signals below the cutoff frequency to pass, while signals above the cutoff frequency are obstructed. 
Thus, we employ typical average pooling as a low-pass filter. However, the cutoff frequencies of different images are different. 
To this end, we control different kernels and strides in multi-groups to generate dynamic low-pass filters. For $m$-th group, we have:
\setlength{\abovedisplayskip}{10pt}
\begin{equation}
\boldsymbol{D}_{m}^{lf}(\boldsymbol{\boldsymbol{v}}^{m}))= \boldsymbol{B}(\Gamma_{s\times s}(\boldsymbol{\boldsymbol{v}}^{m})),
\label{eqn: low-pass}
\end{equation}
where $\boldsymbol{B}(\cdot)$ represents bilinear interpolation and $\Gamma_{s\times s} $ denotes the adaptive average pooling with the output size  of $s \times s$.

\subsubsection{Dynamic High-Pass Filters (DHF)}
High-frequency information is crucial to preserve details in segmentation. As a typical high-pass operator, convolution can filter out irrelevant low-frequency redundant components to retain favorable high-frequency components.
The high-frequency components determine the image quality and the cutoff frequency of the high-pass for each image is different. Thus, we divide the value $\boldsymbol{V}$ into $\boldsymbol{N}$ groups, resulting $\boldsymbol{\boldsymbol{v}}^{n}$. For each group, we use a convolution layer with different kernels to simulate the cutoff frequencies in different high-pass filters. For the $n$-th group, we have:
\setlength{\abovedisplayskip}{10pt}
\begin{equation}
\boldsymbol{D}_{n}^{hf}(\boldsymbol{\boldsymbol{v}}^{n}))=\Lambda_{k\times k}(\boldsymbol{v}^{n}),
\label{eqn: high-pass}
\end{equation}
where $\Lambda_{k\times k}$ denotes the depthwise convolution layer with kernel size of $k\times k$. In addition, we use the Hadamard product of query and high-frequency features to suppress high frequencies inside objects, which are noise for segmentation.

FFN helps to fuse the captured frequency information, but owns a large amount of calculation, which is often ignored in lightweight designs. Here we reduce the dimension of the hidden layer by introducing a convolution layer to make up for the missing capability due to dimension compression. 

\subsubsection{Discuss}
For the frequency similarity kernel, the computational complexity is $\mathcal{O}(hwC^{2})$.
The computational complexity of each dynamic high-pass filter is $ \mathcal{O}(hwCk^{2})$, which is much smaller than that of frequency similarity kernel.
Since the dynamic low-pass filter is implemented by adaptive mean pooling of each group, its computational complexity is about $\mathcal{O}(hwC)$.
Therefore, the computational complexity of a module is linear with the resolution, which is advantageous for high resolution in semantic segmentation.


%

\section{Experiments}\label{sec:Experiments}


%
%

\subsection{Implementation Details}\label{subsection: Implementation Details}
We validate the proposed \ourmodel~on three publicly datasets: ADE20K~\cite{zhou2017scene}, Cityscapes~\cite{cordts2016cityscapes} and COCO-stuff~\cite{caesar2018coco}.
We implement our \ourmodel~with the PyTorch framework base on MMSegmentation toolbox~\cite{mmseg2020}.
Follow previous works~\cite{cheng2021maskformer,zhao2017pyramid}, we use ImageNet-1k to pretrain our model.
%
During semantic segmentation training, we employ the widely used AdamW optimizer for all datasets to update the model parameters. 
%
%
For fair comparisons, our training parameters mainly follow the previous work~\cite{xie2021segformer}.
For the ADE20K and Cityscapes datasets, we adopt the default training iterations 160K in Segformer, where mini-batchsize is set to 16 and 8, respectively. 
For the COCO-stuff dataset, we set the training iterations to 80K and the minibatch to 16.
In addition, we implement data augmentation during training for ADE20K, Cityscapes, COCO-stuff by random horizontal flipping, random resizing with a ratio of 0.5-2.0, and random cropping to $512\times 512$, $1024\times 1024$, $512\times 512$, respectively.
We evaluate the results with mean Intersection over Union (mIoU) metric.

\begin{table}[t!]
\footnotesize 
  \centering
  \caption{Comparison to state of the art methods on ADE20K with resolution at $512 \times 512$. Here we use the Segformer as the baseline and report the percentage growth. MV2=MobileNetV2, EN=EfficientNet, SV2=ShuffleNetV2.}

   {
	\setlength\tabcolsep{4.0pt}
    \begin{tabular}{l|cc|c}
    \hline
    Model & \#Param. & FLOPs & mIoU   \\
    \hline
    FCN-8s &  9.8M     & 39.6G   & 19.7    \\
    PSPNet (MV2) &  13.7M     & 52.2G   & 29.6    \\
    DeepLabV3+ (MV2) &  15.4M     & 25.8G   & 38.1    \\
    DeepLabV3+ (EN) &  17.1M     & 26.9G   & 36.2    \\
    DeepLabV3+ (SV2) &  16.9M     & 15.3G   & 37.6    \\
    Lite-ASPP &  2.9M     & 4.4G   & 36.6    \\
    R-ASPP &  2.2M     & 2.8G   & 32.0    \\
    LR-ASPP &  3.2M     & 2.0G   & 33.1    \\
    HRNet-W18-Small &  4.0M     & 10.2G   & 33.4    \\
    HR-NAS-A &  2.5M     & 1.4G   & 33.2    \\
    HR-NAS-B &  3.9M     & 2.2G   & 34.9    \\
    \hline
    PVT-v2-B0 &  7.6M     & 25.0G   & 37.2    \\
    TopFormer &  5.1M    & 1.8G   & 37.8    \\
    EdgeViT-XXS &  7.9M     & 24.4G   & 39.7    \\
    Segformer (LVT) &  3.9M     & 10.6G   & 39.3    \\
    Swin-tiny &  31.9M     & 46G   & 41.5   \\
    Xcit-T12/16 &  8.4M     & 21.5G   & 38.1    \\
    ViT &  10.2M     & 24.6G   & 37.4   \\
    PVT-tiny &  17.0M     & 33G   & 36.6    \\
    Segformer &  3.8M     & 8.4G   & 37.4   \\
     \hline

    \ourmodel-tiny &  1.6M\footnotesize{\blue{(-58\%)}}     & 2.8G\footnotesize{\blue{(-67\%)}}  & 38.7\footnotesize{\blue{(+1.3)}}  \\ 
    \ourmodel-small &  2.3M\footnotesize{\blue{(-41\%)}}     & 3.6G\footnotesize{\blue{(-61\%)}}  & 40.2\footnotesize{\blue{(+2.8)}}  \\
    \ourmodel-base &  3.0M\footnotesize{\blue{(-21\%)}}     & 4.6G\footnotesize{\blue{(-45\%)}}  & 41.8\footnotesize{\blue{(+4.4)}}   \\
    \hline
    \end{tabular}%
  \label{tab: ADE20K}}
\end{table}%

\begin{table}[t!]
\footnotesize 
  \centering
  \caption{Comparison to state of the art methods on Cityscapes \textit{val} set. The FLOPs are test on the resolution of $1024 \times 2048$. Meanwhile, we also report the percentage increase compared to Segformer.}
   \renewcommand{\arraystretch}{1.0}
   {
	\setlength\tabcolsep{3.5pt}
    \begin{tabular}{l|cc|c}
    \hline
    Model & \#Param. & FLOPs & mIoU     \\
    \hline
    FCN  & 9.8M     & 317G   & 61.5   \\
    PSPNet (MV2)  & 13.7M     & 423G   & 70.2    \\
    DeepLabV3+ (MV2)  & 15.4M     & 555G   & 75.2   \\
    SwiftNetRN  & 11.8M     & 104G   & 75.5   \\
    EncNet  & 55.1M     & 1748G   & 76.9   \\
    Segformer  & 3.8M     & 125G   & 76.2   \\
    \hline
    \ourmodel-tiny  & 1.6M\footnotesize{\blue{(-58\%)}}     & 23.0G\footnotesize{\blue{(-82\%)}}  & 76.5\footnotesize{\blue{(+0.3)}}   \\
    \ourmodel-small  & 2.3M\footnotesize{\blue{(-41\%)}}     & 26.2G\footnotesize{\blue{(-79\%)}}  & 77.6\footnotesize{\blue{(+1.4)}}    \\
    \ourmodel-base  & 3.0M\footnotesize{\blue{(-21\%)}}     & 34.4G\footnotesize{\blue{(-73\%)}}  & 78.7\footnotesize{\blue{(+2.5)}}   \\
    \hline
    \end{tabular}%
  \label{tab: Cityscapes}}%
\end{table}%

\begin{table*}
\footnotesize
    
	\begin{minipage}{0.56\linewidth}
		\caption{Speed-accuracy tradeoffs at different scales on Cityscapes.}
		
        \begin{tabular}{l|cc|c}
    
    \hline
    Model & size  & FLOPs & mIoU   \\
    \hline
    
    Segformer (3.8M) & $512 \times 1024$     & 17.7G   & 71.9   \\
    \ourmodel-base (3.0M) & $512 \times 1024$     & 8.6G\footnotesize{\blue{(-51.4\%)}}  & 73.5\footnotesize{\blue{(+1.6)}}  \\
    \hline
    Segformer (3.8M) & $640 \times 1280$     &  31.5G   & 73.7  \\
    
    \ourmodel-base (3.0M) & $640 \times 1280$ & 13.4G\footnotesize{\blue{(-57.5\%)}}  & 75.6\footnotesize{\blue{(+1.9)}}   \\
    \hline
    Segformer (3.8M) & $768 \times 1536$     & 51.7G   & 75.3   \\
    \ourmodel-base (3.0M) & $768 \times 1536$     & 19.4G\footnotesize{\blue{(-62.5\%)}}  & 76.5\footnotesize{\blue{(+1.2)}}   \\
    \hline
    Segformer (3.8M) & $1024 \times 2048$     & 125G   & 76.2 \\
    \ourmodel-base (3.0M) & $1024 \times 2048$      & 34.4G\footnotesize{\blue{(-72.5\%)}}  & 78.7\footnotesize{\blue{(+2.5)}}   \\
    \hline
    \end{tabular}%
    \label{tab: scale cityscapes}
	    \end{minipage}
	\begin{minipage}{0.35\linewidth}

		  \caption{Comparison to state of the art methods on COCO-stuff. We use a single-scale results at the input resolution of $512 \times 512$. MV3=MobileNetV3}
    \begin{tabular}{l|cc|c}
    \hline
    Model & \#Param. & FLOPs & mIoU    \\
    \hline

    PSPNet (MV2)  & 13.7M     & 52.9G   & 30.1   \\
    DeepLabV3+ (MV2)  & 15.4M     & 25.9G   & 29.9    \\
    DeepLabV3+ (EN)  & 17.1M     & 27.1G   & 31.5   \\
    LR-ASPP (MV3)  & --     & 2.37G   & 25.2  \\
    \hline
    \ourmodel-base & 3.0M     & 4.6G  & 35.1  \\
    \hline
    \end{tabular}%
    \label{tab: COCO-stuff}
	\end{minipage}
\end{table*}

\subsection{Comparisons with Existing Works}

\subsubsection{Results on ADE20K Dataset.}
We compare our \ourmodel~with top-ranking semantic segmentation methods, including CNN-based and vision Transformer-based models. 
Following the inference settings in~\cite{xie2021segformer}, we test FLOPs at $512 \times 512$ resolution and show the single scale results in Table~\ref{tab: ADE20K}. 
Our model \ourmodel-base improves by 5.2 mIoU under the same computing power consumption as Lite-ASPP, reaching 41.8 mIoU.
At the same time, by reducing the number of layers and channels, we obtain \ourmodel-tiny and \ourmodel-small versions to adapt to different computing power scenarios.
For the lightweight and efficient Segformer (8.4 GFLOPs),our base version (4.6 GFLOPs) also gain 4.4 mIoU using half the computing power and the tiny version (2.4 GFLOPs) with only 1/4 the computing power improving 1.3 mIoU.
Only 1.8 GFLOPs are needed for the lighter topformer, but our base version has 2.1M less parameters (5.1M vs 3M) with 4.0 higher mIoU.

\begin{table}[t!]
\small
  \centering

  \caption{Ablation studies on the parallel structure.}
   \renewcommand{\arraystretch}{1.0}
   {
	\setlength\tabcolsep{6pt}
    \begin{tabular}{cccc}
    \hline
    Setting   & \#Param. & FLOPs &  mIoU\\
    \hline
    w/o PD & 2.78G & 2.98G & 39.2\\
    w/o PL  & 0.42G & 1.65G & 19.5\\
    Parallel & 3.0G & 4.6G & 41.8\\
    \hline
    \end{tabular}%
  \label{tab: Parallel}}%
\end{table}%

\begin{table}[t!]
\small
  \centering

  \caption{Ablation studies on heterogeneous architecture. }

   \renewcommand{\arraystretch}{1.0}
   {
	\setlength\tabcolsep{6pt}
    \begin{tabular}{cccc}
    \hline
    Setting  &  \#Param.  & FLOPs  &mIoU\\
    \hline
    All PD  & 0.6M & 1.85G & 27.4\\
    All PL  & 3.6M & 7.0G & 41.6\\
    Heterogeneous & 3.0M & 4.6G & 41.8\\
    \hline
    \end{tabular}%
  \label{tab:Heterogeneous}}%
\end{table}%

\subsubsection{Results on Cityscapes Dataset.}
Table~\ref{tab: Cityscapes} shows the results of our model and the cutting-edge methods on Cityscapes.
Although the Segformer is efficient enough, due to its square-level complexity, we only use 30\% of the computational cost to reach 78.7 mIoU, which is 2.5 mIoU improvement with a 70\% reduction in FLOPs. 
Meanwhile, we report the results at different high resolutions in Table~\ref{tab: scale cityscapes}. At the short side of \{512, 640, 768, 1024\}, the computational cost of our model is 51.4\%, 57.5\%, 62.5\% and 72.5\% of that of Segformer, respectively. 
Meanwhile, the mIoU are improved by 1.6, 1.9, 1.2 and 2.5, respectively. 
The higher the input resolution, the more advantageous of our model  in both computational cost and accuracy. 

\subsubsection{Results on COCO-stuff Dataset.}
COCO-stuff dataset contains a large number of difficult samples that collected in COCO. 
As show in Table~\ref{tab: COCO-stuff}, although complex decoders (\eg, PSPNet, DeepLabV3+) can achieve better results than LR-ASPP (MV3), they bring a lot of computational cost. Our model achieves an accuracy of 35.1 mIoU while only taking 4.5 GFLOPs, achieving the best trade-off.

\subsection{Ablation Studies}
All the ablation studies are conducted on ADE20K dataset with \ourmodel-base unless otherwise specified.

\subsubsection{Rationalization of Parallel Structures.}
Parallel architecture is the key to removing the decoder head and ensuring accuracy and efficiency. 
We first adjust the proposed structure to a naive pyramid architecture (denoted as ``w/o PD") and a ViT architecture (denoted as ``w/o PL") to illustrate the advantages of the parallel architecture. Specifically, the ``w/o PD" means removing PD module and keeping only PL module, while the ``w/o PL" does the opposite.
As shown in Table~\ref{tab: Parallel}, the setting ``w/o PD" reduces 2.6 mIoU due to the lack of high-resolution pixel semantic information. 
The ``w/o PL" structure without the pyramid structure has a significant reduction in accuracy due to few parameters and lack of rich image semantic information.
It also demonstrates that our parallel architecture can effectively combine the advantages of both architectures.

\subsubsection{Advantages of Heterogeneous Structure.}
The purpose of the heterogeneous approach is to further reduce the computational overhead.
The PL module is adopted to learn the prototype representation in the clustered features, and then use PD to combine the original features for restoration, which avoids direct calculation on the high-resolution original features and reduce the computational cost. 
It can be seen from Table~\ref{tab:Heterogeneous} that when the parallel branch is adjusted to the pixel description module (denote as ``All PD"), which means that the prototype representation is learned by PD module. 
The model size is only 0.6M, and the FLOPs are reduced by 2.5G, but the accuracy is reduced by 14.3 mIoU. This is due to the PD lacks the ability to learn great prototype representations.
In contrast, after we replace the PD module with the PL module (denote as ``All PL"), the FLOPs are increased by 2.4G, but there is almost no difference in accuracy. We believe that the PD module is actually only a simple way to restore the learned prototype, and the relatively complex PL module saturates the model capacity. 

\subsubsection{Advantages of Adaptive Frequency Filter.}
We use two datasets with large differences, including ADE20K and Cityscapes, to explore the core components in adaptive frequency filter module. 
The main reason is that the upper limit of the ADE20K dataset is only 40 mIoU, while the upper limit of the Cityscapes is 80 mIoU. 
The two datasets have different degrees of sensitivity to different frequencies.
We report the benefits of each internal component in the Table~\ref{tab:operator_combination}.
We find that DHF alone outperforms DLF, especially on the Cityscapes dataset by 2.6 mIoU, while FSK is significantly higher than DLF and DHF on ADE20K. 
This shows that ADE20K may be more inclined to an intermediate state between high frequency and low frequency, while Cityscapes needs more high frequency information.
The combined experiments show that the combination of the advantages of each component can stably improve the results of ADE20K and Cityscapes.
\subsubsection{Frequency Statistics Visualization.}
\begin{figure}[t!]
	\centering
    \small
	\begin{overpic}[width=0.95\linewidth]{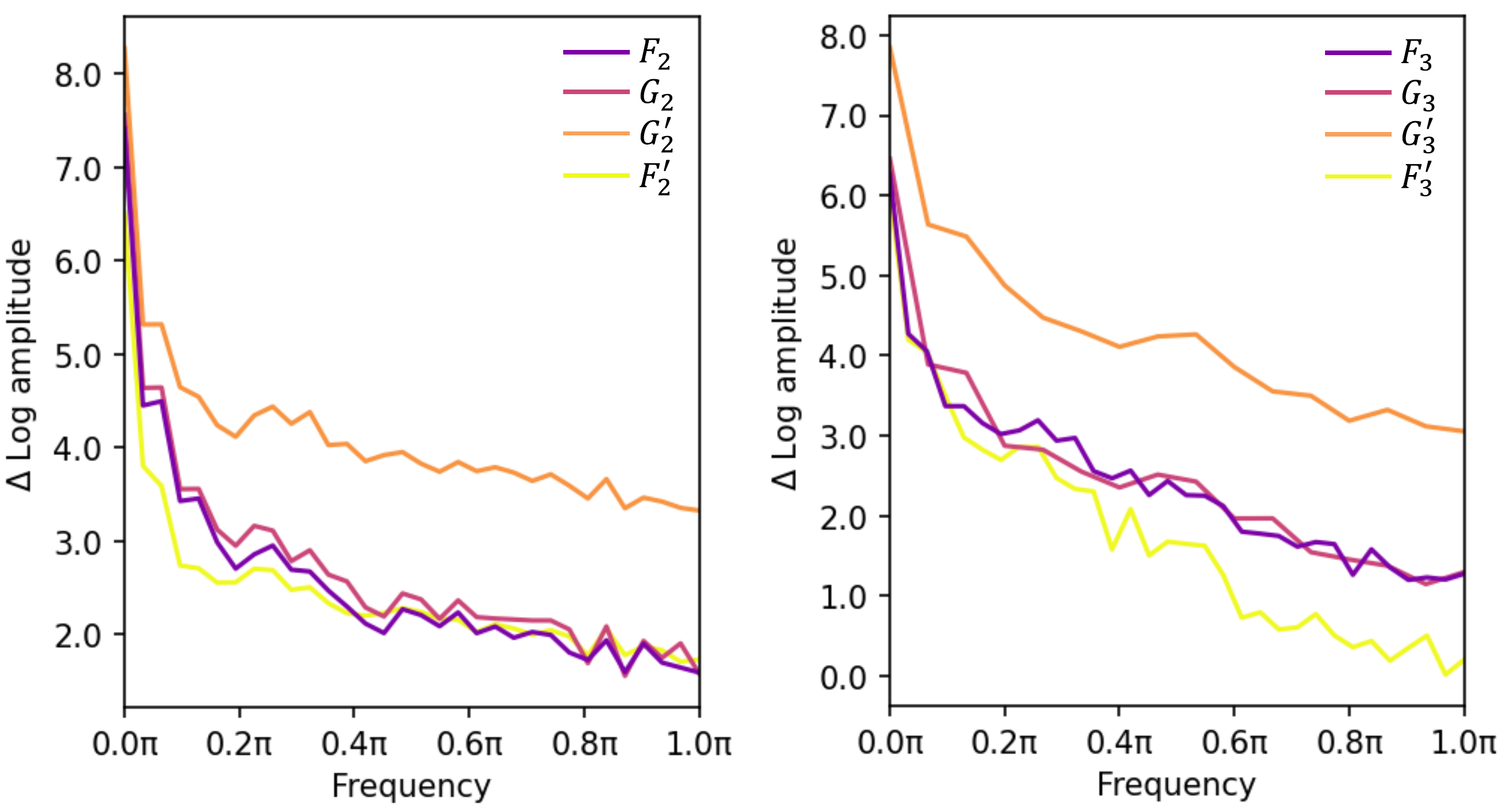}

    \end{overpic}
	\caption{Frequency analysis of stage-2 (left) and stage-3 (right).}
    \label{figure: fq1}
\end{figure}

\begin{figure}[t!]
	\centering
    \small
	\begin{overpic}[width=0.95\linewidth]{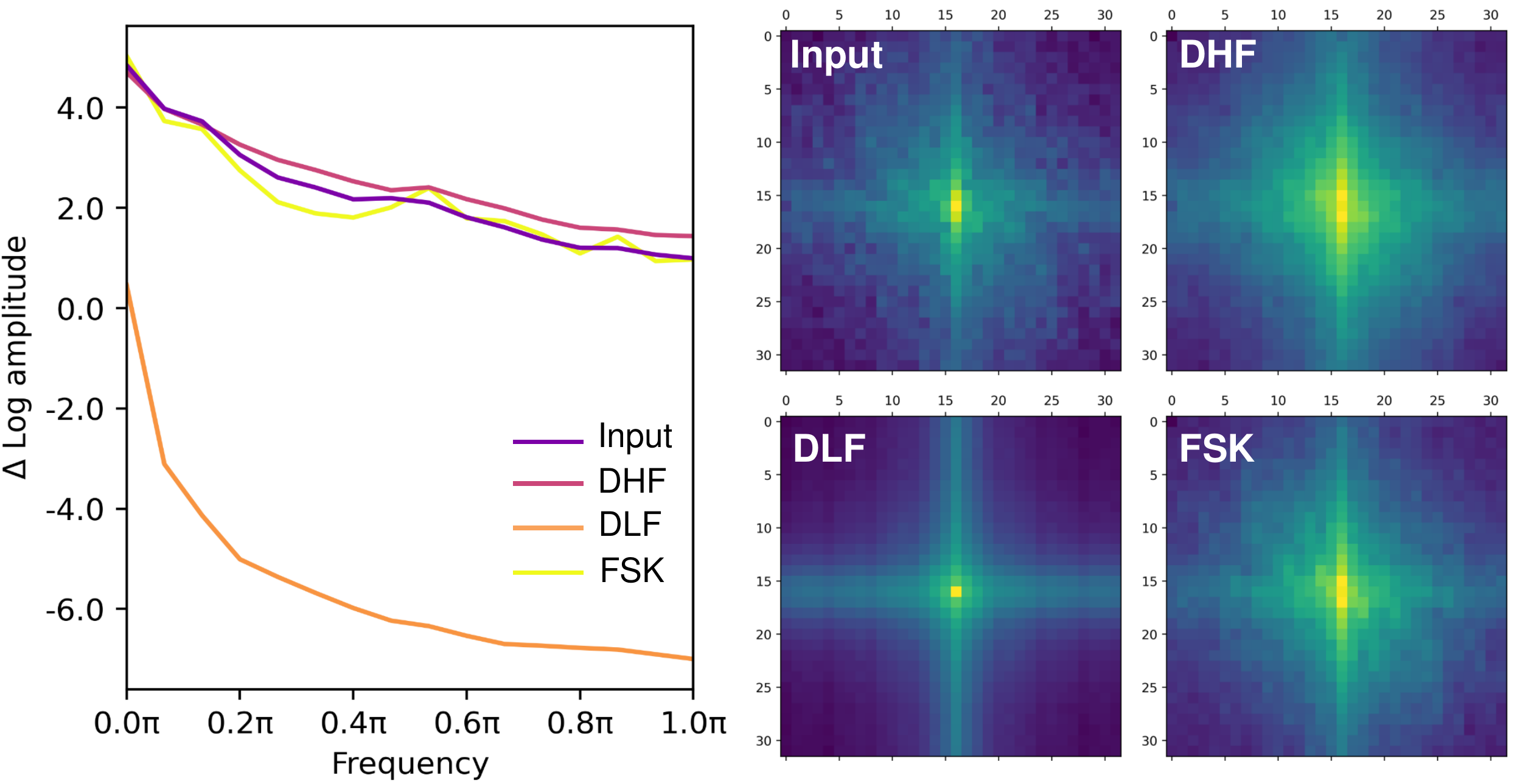}
    \end{overpic}
	\caption{Frequency analysis of the core components in PL module.}
    \label{figure: fq2}
\end{figure}

We first count the characteristic frequency distribution of different stages, as shown in Figure~\ref{figure: fq1}. 
It can be found that the curves of $G_2$ and $F_2$ almost overlap, indicating that the frequencies after clustering are very similar to those in the original features. The same goes for $G_3$ and $F_3$. 
Whereas, the learned prototype representation after frequency adaptive filtering significantly improves the contained frequency information. After PD restoration, different frequency components can be emphasized in different stages.
As shwon in Figure~\ref{figure: fq2}, we also analyze the frequency effects of the core components in the AFF module. As expected, DLF and DHF show strong low-pass and high-pass capabilities, respectively, as FSK does. At the same time, we also found that the important frequency components screened and enhanced by FSK are mainly concentrated in the high frequency part, but the frequency signal is more saturated than that of DHF. This also shows that the high-frequency component part is particularly important in the semantic segmentation task, because it emphasizes more on the boundary details and texture differences between objects. Meanwhile, according to the analysis in Table~\ref{tab:operator_combination} (the effects of ADE20K and Cityscapes have been steadily improved), each core component has its own advantages, and the AFF module shows strong robustness in various types and complex scenes.
\subsubsection{Speed and Memory Costs.}
Meanwhile, we report the speed on the Cityscapes dataset in Table~\ref{tab: fps}. We can find that the proposed model improves by 10 FPS and performs much better than Segformer on such high-resolution Cityscapes images.

\begin{table}[t!]
\small
  \centering

  \caption{Ablation studies on frequency aware statistics.}
   \renewcommand{\arraystretch}{1.0}
   {
	\setlength\tabcolsep{4pt}
    \begin{tabular}{lcccc}
    \hline
    Setting  & \#Param.  & FLOPs & ADE20K & Cityscapes \\
    \hline
    DLF & 2.4M & 3.6G & 38.7 & 75.7\\
    DHF & 2.6M & 3.9G & 39.3 & 78.3\\
    FSK & 2.9M & 4.2G & 40.5 & 75.3\\
    DLF + DHF & 2.7M & 3.9G & 41.1 & 77.8 \\
    DLF + FSK & 2.8M & 4.2G & 40.0 & 76.2\\
    DHF + FSK & 2.9M & 4.3G & 41.2 & 77.3 \\
    Whole & 3.0M & 4.6G & 41.8 &78.7 \\

    \hline
    \end{tabular}%
  \label{tab:operator_combination}}%
\end{table}%
\begin{table}[t!]
\small
  \centering

  \caption{The FPS is tested on a V100 NVIDIA GPU with a batch size of 1 on the resolution of 1024x2048.}
   \renewcommand{\arraystretch}{1.0}
   {
	\setlength\tabcolsep{6pt}
    \begin{tabular}{cccc}
    \hline
    Model   & FPS &  mIoU\\
    \hline
    Segformer & 12 & 76.2\\
    \ourmodel   & 22 & 78.7\\

    \hline
    \end{tabular}%
  \label{tab: fps}}%
\end{table}%

\section{Conclusion} 
In this paper, we propose \ourmodel, a head-free lightweight semantic segmentation specific architecture.
The core is to learn the local description representation of the clustered prototypes from the frequency perspective, instead of directly learning all the pixel embedding features.
It removes the complicated decoder while having linear complexity Transformer and realizes semantic segmentation as simple as regular classification. 
The various experiments demonstrate that the \ourmodel~owns powerful accuracy and great stability and robustness at low computational cost. 
%

\section{Acknowledgements} This work was supported by Alibaba Group through Alibaba Research Intern Program.
\bibliography{aaai23}

\end{document}